# Semi-Supervised Anomaly Detection Pipeline for SOZ Localization Using Ictal-Related Chirp

Nooshin Bahador[1], Milad Lankarany[1,2,3,4]

[1] Krembil Research Institute – University Health Network (UHN), Toronto, ON, Canada
[2] Institute of Biomaterials & Biomedical Engineering (IBBME), University of Toronto, Toronto, ON, Canada
[3] Department of Physiology, University of Toronto, Toronto, ON, Canada
[4] KITE Research Institute, Toronto Rehabilitation Institute - University Health Network (UHN), Toronto, ON, Canada

* To whom correspondence should be addressed:
Milad Lankarany (milad.lankarany@uhn.ca)
The Krembil Research Institute – University Health Network (UHN)
60 Leonard Ave, Toronto, M5T 0S8, Canada

**Abstract**

This study presents a quantitative framework for evaluating the spatial concordance between clinically defined seizure onset zones (SOZs) and statistically anomalous channels identified through time-frequency analysis of chirp events. The proposed pipeline employs a two-step methodology: (1) Unsupervised Outlier Detection, where Local Outlier Factor (LOF) analysis with adaptive neighborhood selection identifies anomalous channels based on spectro-temporal features of chirp (Onset frequency, offset frequency, and temporal duration); and (2) Spatial Correlation Analysis, which computes both exact co-occurrence metrics and weighted index similarity, incorporating hemispheric congruence and electrode proximity. Key findings demonstrate that the LOF-based approach ($n_{neighbors}$=20, contamination=0.2) effectively detects outliers, with index matching (weighted by channel proximity) outperforming exact matching in SOZ localization. Performance metrics (precision, recall, F1) were highest for seizure-free patients (Index Precision: 0.903 ± 0.168) and those with successful surgical outcomes (Index Precision: 0.865 ± 0.157), whereas failure cases exhibited lower concordance (Index Precision: 0.460 ± 0.456). The key takeaway is that chirp-based outlier detection, combined with weighted spatial metrics, provides a complementary method for SOZ localization, particularly in patients with successful surgical outcomes.

# 1. Introduction

Epileptic seizures arise from sudden, uncontrolled surges of aberrant electrical discharges in the brain, primarily due to disrupted equilibrium between neural excitation and inhibition. These disturbances heighten neuronal excitability, impairing standard cognitive and physiological processes. When seizures recur spontaneously, the condition progresses to epilepsy—a persistent neurological disorder with diverse clinical manifestations. The etiology of seizures is complex, encompassing factors such as oxygen deprivation, genetic abnormalities, structural brain malformations, and systemic illnesses. Symptoms vary widely, impacting movement, consciousness, and sensory perception, posing challenges for both diagnosis and treatment (Freeman et al., 1993). A critical method for detecting epileptic activity is electroencephalography (EEG), which categorizes abnormal discharges into three phases: ictal (active seizure), interictal (between seizures), and postictal (recovery period). Research has extensively examined the transition mechanisms within the ictal phase, including onset, spread, and cessation (e.g., Miri et al., 2018; Rich et al., 2020). Studies suggest that seizure initiation often involves inhibitory synchronization, leading to reduced neuronal firing and increased extracellular potassium concentrations (de Curtis and Avoli, 2016). These electrophysiological shifts generate unique time-frequency signatures, such as "chirps"—transient signals characterized by progressive frequency modulations within a specific band (Grinenko et al., 2018). In intracranial EEG studies, chirp-like patterns have been frequently documented during epileptic episodes (Bahador et al., 2024; Bahador et al., 2025; Li et al., 2020; Gnatkovsky et al., 2011; Kurbatova et al., 2016; Sen et al., 2007; Niederhauser et al., 2003; Schiff et al., 2000; Feltane et al., 2013; Gnatkovsky et al., 2019a; Benedetto and Colella, 1995). Emerging evidence indicates that high-frequency chirps may help pinpoint epileptogenic regions (Di Giacomo et al., 2024), though their timing remains debated—appearing either prior to seizure onset or during propagation (e.g., Kurbatova et al., 2016; Niederhauser et al., 2003). Recent findings suggest that surgical removal of brain areas producing these chirps correlates with improved outcomes, reinforcing their utility in clinical decision-making. Di Giacomo et al. (2024) demonstrated that fast oscillatory "chirps" in stereoelectroencephalography (SEEG) recordings reliably identify the epileptogenic zone in drug-resistant focal epilepsy. These dynamic high-frequency patterns were consistently observed across various seizure types and anatomical locations, aligning with clinically determined seizure foci. Notably, resection of chirp-associated regions led to better postoperative results, supporting the use of chirp analysis as an objective method for refining localization, especially in cases with ambiguous imaging findings.

Despite growing evidence that ictal chirps may serve as biomarkers for SOZs and surgical outcomes, critical gaps persist in understanding their spatiotemporal behavior and translational relevance. In this study, we leverage intracranial EEG (iEEG) recordings from a multicenter cohort to analyze ictal chirp patterns through a hybrid annotation pipeline, combining manual expert review with automated ridge detection for feature extraction. We develop a quantitative framework to evaluate the spatial concordance between chirp-derived outliers and clinically defined seizure onset zones (SOZs), employing unsupervised outlier detection (Local Outlier Factor) and novel spatial correlation metrics (exact and weighted index matching). Further, we transform chirp characteristics into interpretable 3D and radial embeddings to visualize their discriminative power in SOZ localization. By correlating these data-driven findings with surgical outcomes, we provide evidence that chirp-based biomarkers not only align with clinically annotated epileptogenic tissue but also hold predictive value for postoperative success.

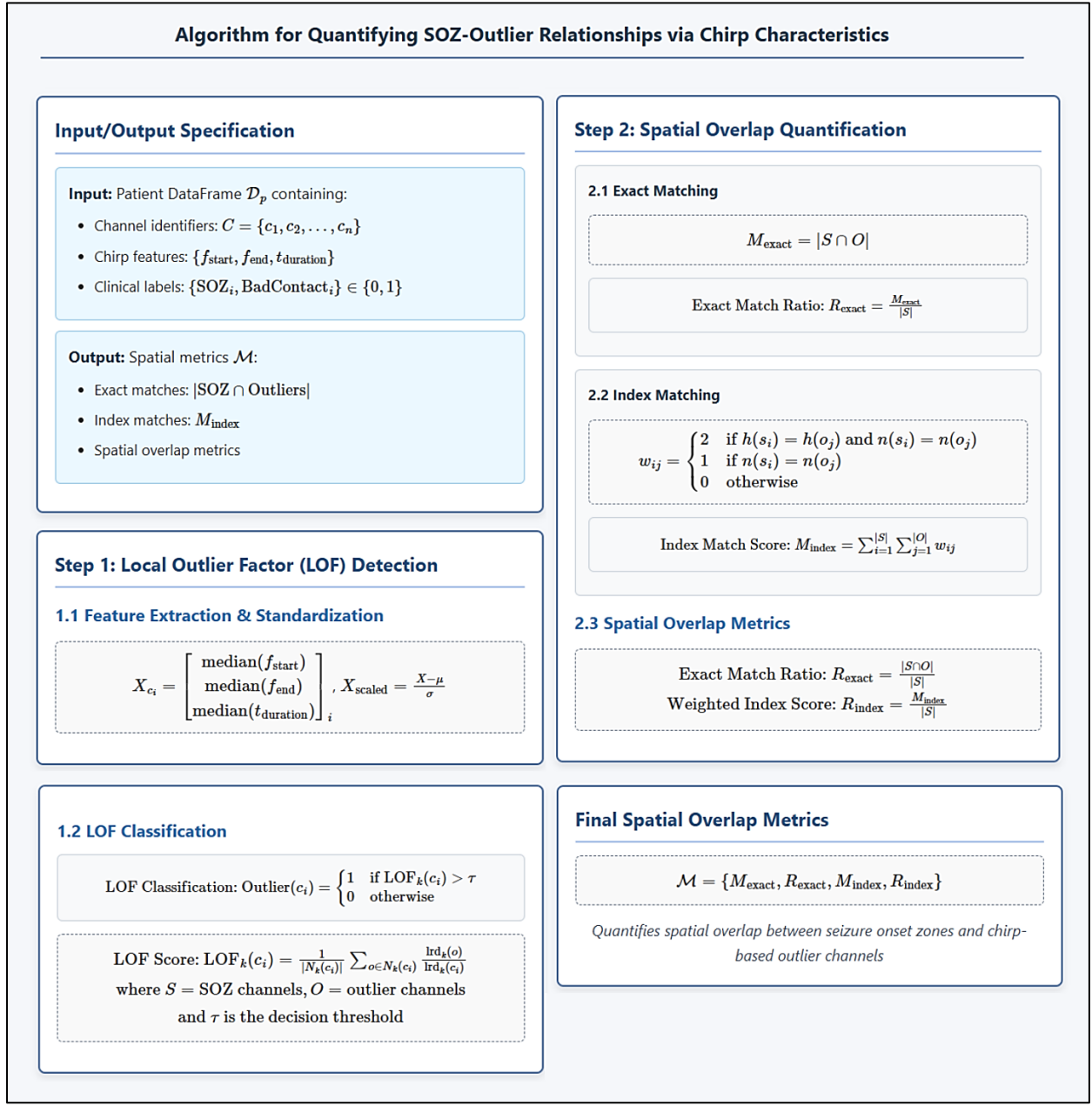

**Fig.1. Quantitative Framework for Spatial Co-localization of Ictal Onset Zones and Spectral Outliers via Chirp-based Feature Analysis:** This algorithm quantifies the relationship between seizure onset zones (SOZ) and outlier channels using chirp characteristics through a two-step

process. First, it identifies outlier channels by analyzing three key chirp features (start frequency, end frequency, and duration) for each channel, standardizing these values, then applying a Local Outlier Factor analysis that flags channels as outliers when their abnormality score exceeds a set threshold. Second, it measures spatial overlap by calculating both exact matches (counting channels that are both SOZ and outliers) and index matches (assigning weighted scores based on electrode numbering and hemisphere location matches). The final output includes four key metrics: the count of exact matches, the exact match ratio, the weighted index score, and the index match ratio - collectively providing a comprehensive assessment of how well the detected outliers align spatially with clinically identified seizure onset zones.

## 2. Method

### 2.1. Annotated iEEG Spectrogram Dataset for Chirp Pattern Analysis

The study utilized intracranial EEG (iEEG) data from the Epilepsy-iEEG-Multicenter-Dataset (hosted by OpenNeuro), comprising recordings from 13 patients across four centers, stored in BIDS format with clinical metadata (Li et al., 2022). A derivative dataset of 22,721 spectrograms was generated, where chirp patterns were semi-automatically annotated using a hybrid approach: users manually drew bounding boxes, and an algorithm performed ridge detection, fitting models (ex. linear/exponential) while extracting spectro-temporal features (Onset/offset time/frequency, temporal/spectral duration, RMSE, R², direction). Clinical features included seizure onset zone, epileptogenic region, surgical outcomes (Engel/ILAE scores), and case difficulty. The annotated dataset and supplementary materials were made publicly available on GitHub.

### 2.2. Pipeline for Spatial Concordance of Spectral Outliers and Seizure Onset Zones

The proposed pipeline shown in Fig.1 evaluates the spatial concordance between clinically defined seizure onset zones (SOZs) and statistically anomalous channels identified through their time-frequency characteristics. The methodology employs: I) Unsupervised Outlier Detection: 1- Extracts tri-dimensional feature vectors (median start frequency, end frequency, temporal duration) from chirp events. 2- Normalizes features via z-scoring ($\mu = 0, \sigma = 1$). 3- Implements Local Outlier Factor (LOF) analysis with adaptive neighborhood selection (k-nearest neighbors). 4- Classifies outliers using density-based thresholding ($\tau$) on LOF scores. II) Spatial Correlation Analysis: 1- Computes exact co-occurrence metrics (intersection cardinality). 2- Derives weighted index similarity accounting for: Hemispheric congruence ($h(s_i) = h(o_j)$) and Electrode numbering proximity ($n(s_i) = n(o_j)$). 3- Generates normalized overlap indices (range [0,1]). Output metrics ($M_{exact}$, $R_{exact}$, $M_{index}$, $R_{index}$) provide complementary measures of spatial correspondence, enabling evaluation of the hypothesis that spectral outliers colocalize with ictogenic tissue. The framework's mathematical formulation ensures sensitivity to both spatial precision (exact matches) and regional proximity (weighted indices). While Exact Match requires the SOZ and outlier channel names to be identical, Index Match allows for partial matches based on two criteria: (1) the first character of the channel name (e.g., 'MST1' → 'M'), and (2) the numeric part of the name (e.g., 'MST1' → '1').

A fixed contamination parameter of 0.2 was employed in the Local Outlier Factor (LOF) algorithm to identify anomalous data channels. This configuration corresponds to selecting the top 20% of data points with the highest LOF scores as outliers. Unlike threshold-based methods, the LOF algorithm operates by first computing a local density-based anomaly score for each point. The procedure involves three primary steps:

(a) computing LOF scores for all points, where higher scores indicate greater deviation from the local neighborhood density.
(b) ranking all points according to their LOF scores.
(c) selecting the top 20% of points as outliers, as determined by the contamination parameter.

It is important to note that the LOF score threshold used to delineate outliers is not fixed across all samples. Rather, it is adaptively determined per patient, as the underlying data distribution varies across individuals. Consequently, while the proportion of flagged outliers remains constant at 20% across patients, the specific LOF score corresponding to this percentile differs. To account for inter-patient variability, all features were standardized on a per-patient basis prior to LOF computation. This ensures that LOF scores are computed relative to the distribution of each individual patient's data, thus maintaining consistency in anomaly detection sensitivity.

## 2.3. Visualizing Chirp Features for SOZ Localization and Outcome Prediction

To further demonstrate how spectro-temporal chirp characteristics aid in seizure onset zone (SOZ) localization and surgical outcome prediction in epilepsy patients, chirp annotations were converted into multi-dimensional feature representations. This involved extracting temporal and spectral features, projecting them into polar coordinate visualizations, and applying channel-specific color mappings for enhanced interpretability. Fig.2 illustrates the methodology for generating feature embeddings, including preprocessing, temporal/spectral feature extraction, HSL-based channel coloring, and the creation of 2D/3D scatter plots and clinical radar charts. Fig.3 details the radial visualization pipeline, where temporal duration, frequency bandwidth, and chirp slope are transformed into polar coordinates, angular distributions represent feature types, while scaled radial distances encode magnitudes, excluding poor-quality recordings for robust analysis. Together, these visualizations provide an intuitive framework for chirp-based SOZ assessment.

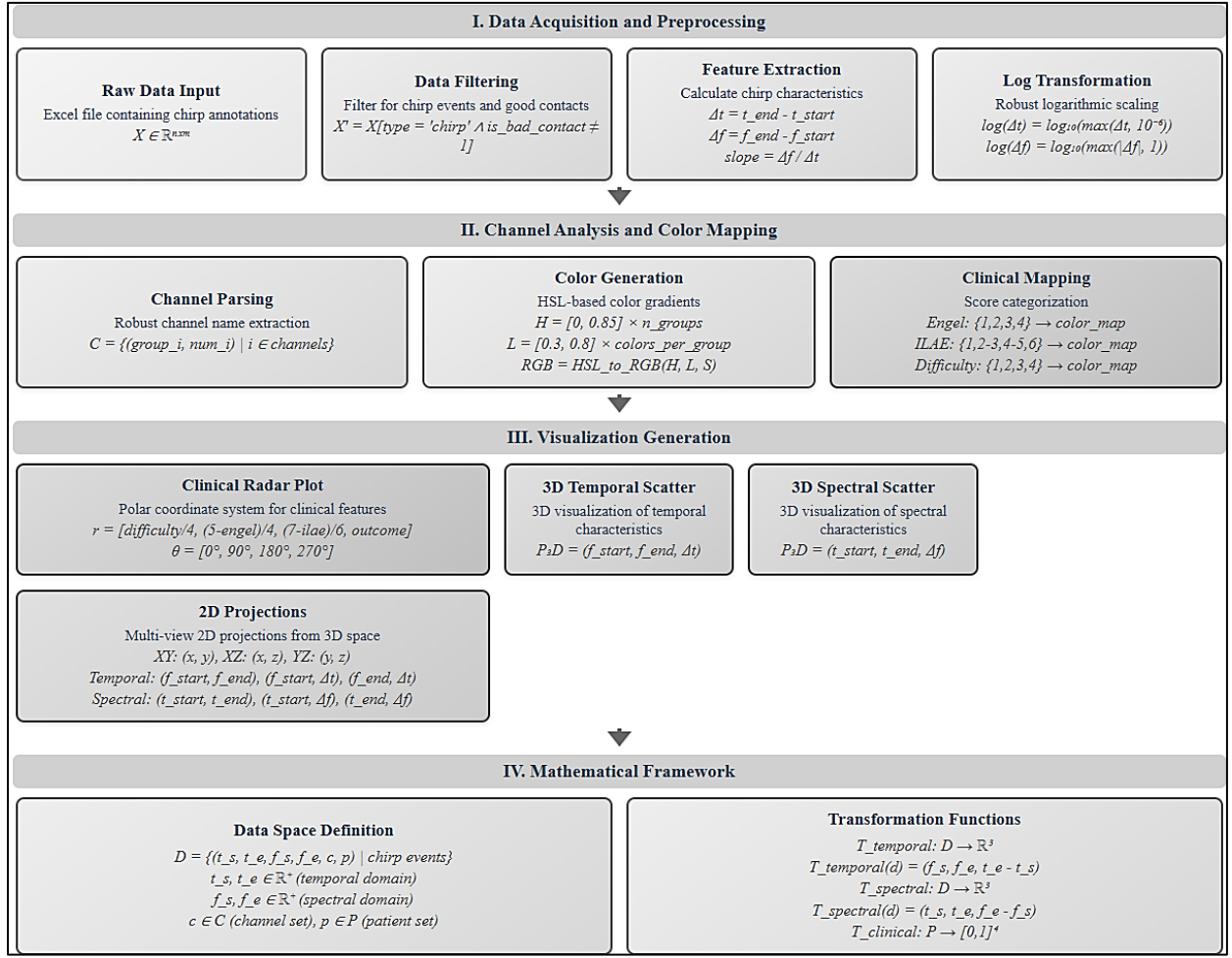

**Fig.2. Methodology for generating Feature Embedding and extracting 2D/3D scatter plots and clinical radar charts from chirp annotation data**, including data preprocessing, feature extraction (temporal and spectral durations), HSL-based channel color mapping, and multi-dimensional visualization generation with polar coordinate clinical profiling.

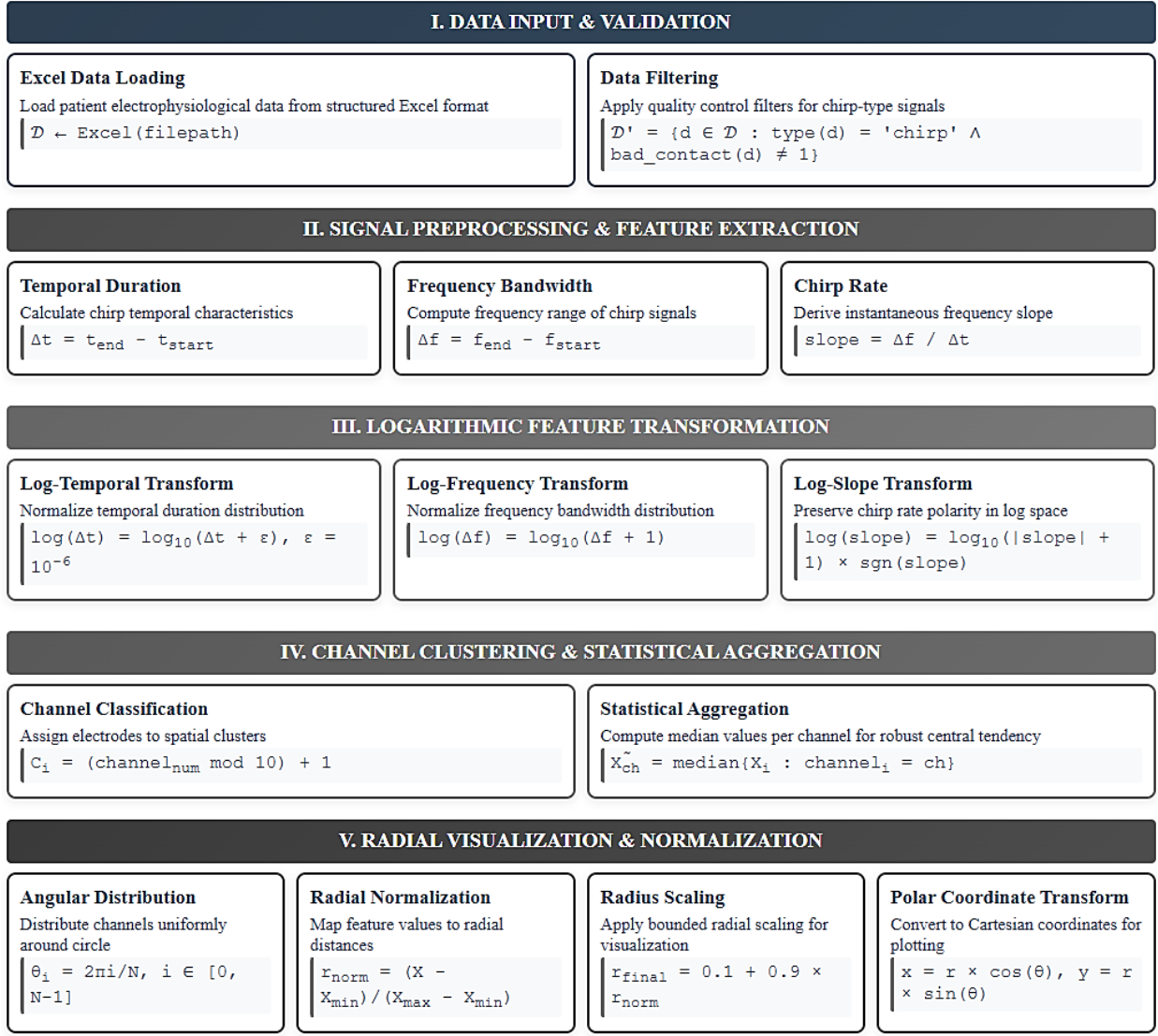

**Fig.3. Methodology for Radial Visualization from the chirp annotation data:** This figure illustrates the analytical pipeline for processing chirp-type events while excluding poor contact recordings. The visualization pipeline transforms temporal duration (Δt), frequency bandwidth (Δf), and instantaneous chirp rate (slope) into polar coordinate systems through angular distribution and radial normalization, mapping feature magnitudes to scaled radial distances for intuitive interpretation.

## 3. Results

Given existing evidence in the literature suggesting that ictal chirps may serve as complementary markers of seizure onset zone (SOZ) channels, we hypothesized that SOZ-associated chirps possess specific spectro-temporal characteristics that distinguish them from non-SOZ channels. Specifically, we proposed that channels identified as spatial outliers within a chirp-based feature space would show strong concordance with SOZ contacts annotated by the clinical team. These outlier channels are expected to occupy distinct regions of the feature space, deviating from the main clusters of non-SOZ channels. Therefore, we argue that chirp-based feature outliers are not random anomalies, but are instead indicative of underlying epileptogenic regions. To test this hypothesis, we implemented our proposed pipeline for detecting, mapping, and visualizing outliers within the chirp feature space.

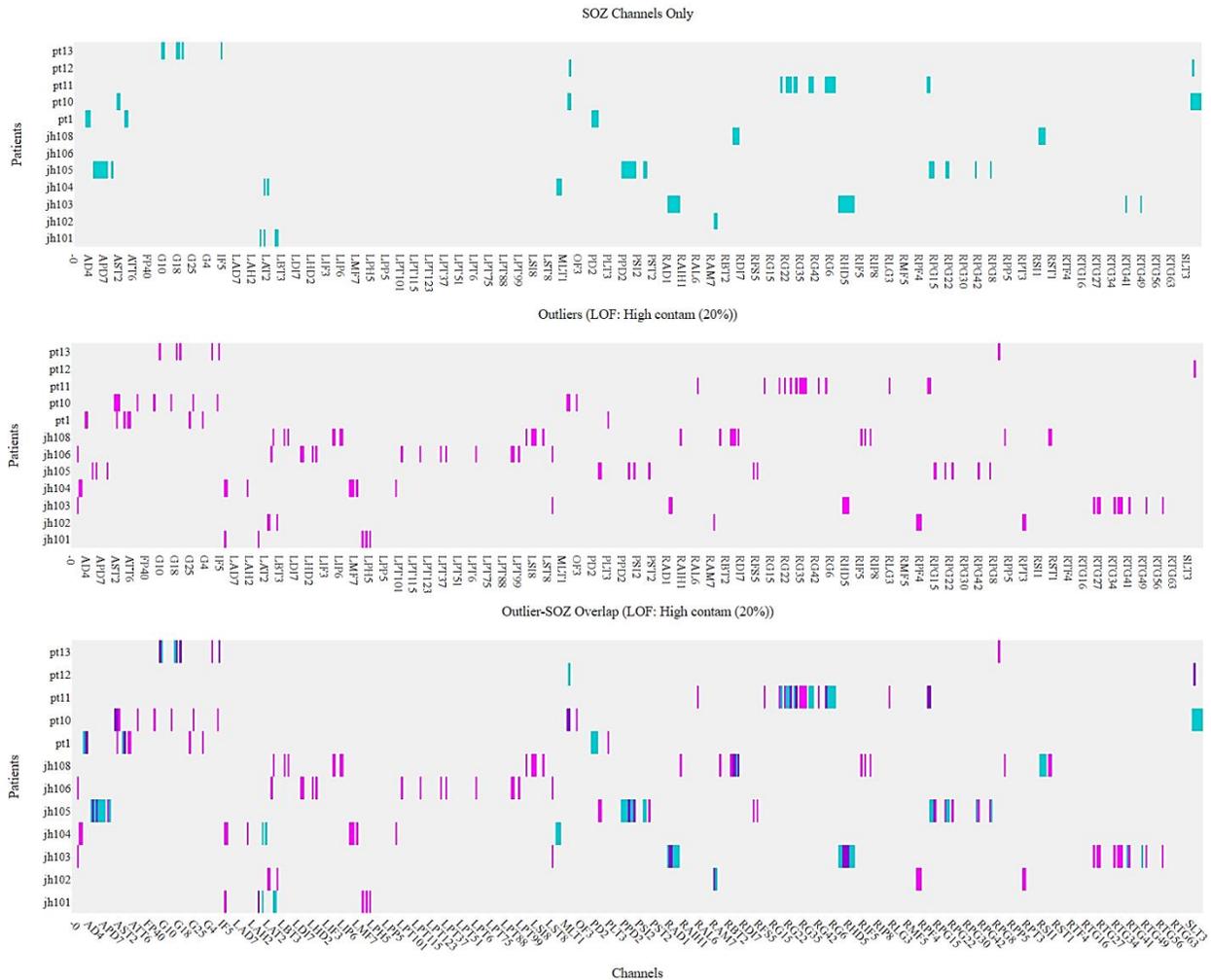

**Fig.4. Comparison of SOZ Channels and LOF-Detected Outliers (20% Contamination):** LOF with $n_{neighbors} = 20$ and explicit 20% contamination identified outliers from median spectral-temporal features. Upper: Clinical SOZ annotations (teal). Middle: LOF outliers (magenta) forced to 20% prevalence. Lower: Overlap states (dark violet=both, teal=SOZ-only, magenta=outlier-only, gray=neither) with value-mapped colorscale. 3D embeddings for visual confirmation are provided in the Appendix. Specifically, the 3D feature embeddings (Fig. 4 (Supp1)–Fig. 4 (Supp4)) and radial projections (Fig. 4 (Supp5)–Fig. 4 (Supp7)) further validated the relationship between chirp outliers and SOZ annotations. In several patients (e.g., jh101, pt13, pt1), channels associated with the SOZ consistently appeared as distinct spatial outliers. These visualizations offer an intuitive complement to the statistical analysis.

The Local Outlier Factor (LOF) algorithm with $n_{neighbors} = 20$ and $contamination = 0.2$ was applied to patient chirp data, where median channel features ($start\ freq, end\ freq, duration\ time$) were standardized using scikit-learn's StandardScaler (Pedregosa *et al.*, 2011). Outliers were identified as the top 20% most anomalous points ($contamination = 0.2$) via LOF's prediction method. Fig.4 comprises three heatmaps: (1) Upper SOZ annotations (teal=SOZ, gray=non-SOZ), (2) Middle LOF outliers (magenta=predicted outliers at 20% contamination), and (3) Lower overlap analysis using discrete thresholds (dark violet=SOZ+outlier, teal=SOZ-only, magenta=outlier-only, gray=neither).

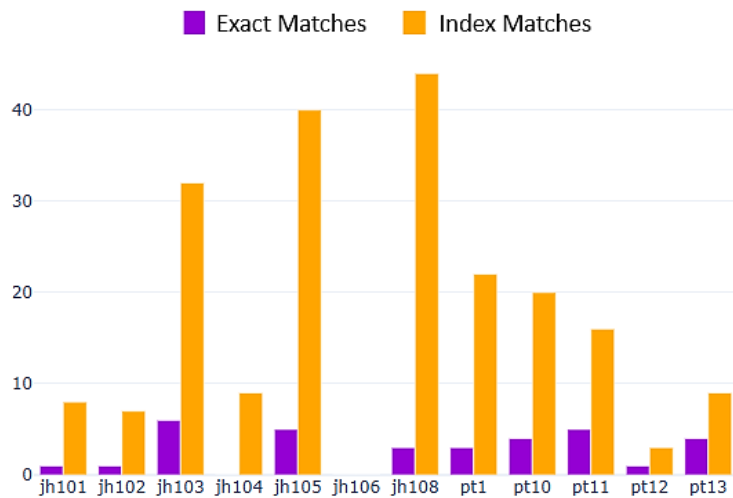

**Fig.5. Spatial Overlap Metrics Comparing Exact and Index Matches:** Bar plot showing two spatial matching metrics between SOZ (seizure onset zone) and outlier channels across multiple patients. The y-axis displays absolute counts for two distinct spatial overlap metrics. The Exact Matches (dark violet bars) represent the raw count of channels classified as both SOZ and outliers, meaning this value can never exceed the total number of SOZ channels for a given patient. In contrast, the Index Matches (orange bars) reflect a weighted scoring system that accounts for electrode label similarities: matches earn 2 points if an SOZ and outlier channel share the same electrode number and hemisphere (indicated by the same initial character), or 1 point if they share only the electrode number but differ in hemisphere. Because the Index Match score sums all possible pairwise comparisons—and doubles same-hemisphere matches—it can substantially exceed the patient's total SOZ count. For example, a patient with 5 SOZ channels could theoretically reach a maximum score of 10 (2 points × 5 channels) if every SOZ channel perfectly matches an outlier in both number and hemisphere. Thus, while Exact Matches indicate direct

overlap, Index Matches quantify the broader spatial relationship. Actual SOZ channel counts — jh101: 4, jh102: 2, jh103: 18, jh104: 5, jh105: 26, jh106: 0, jh108: 8, pt1: 9, pt10: 10, pt11: 17, pt12: 2, pt13: 6.

The Local Outlier Factor (LOF) Detection was performed using a neighborhood size of n_neighbors=20 and a fixed contamination rate of 20%, applied to median spectral-temporal features (start frequency, end frequency, and duration) after standardization. Following outlier identification, the Spatial Metric Calculation Pipeline quantified spatial relationships between outliers and seizure onset zone (SOZ) channels. Exact Matching counted channels classified as both SOZ and outliers, while a weighted scoring system was applied during index matching, giving higher priority to electrode labels that matched both the initial character and numerical digits.with higher weights (2x) for initial character matches. These metrics were visualized in Fig.5 showing both raw counts of exact matches and weighted index matches across patients, providing complementary measures of special convergence between outlier channels and clinically defined SOZ regions. Exact Match only captures perfect electrode overlaps with the SOZ (Seizure Onset Zone), which means it misses nearby electrodes or those with matching numbers that may still be clinically relevant. In contrast, Index Match provides a more nuanced measure by capturing how close mismatched electrodes are to the SOZ, allowing quantification of spatial spread—such as pathological activity in neighboring or anatomically related regions. For example: Electrodes in the same brain region (same hemisphere and electrode number) receive +2 points. Electrodes in homologous regions (same number but opposite hemisphere, reflecting hemispheric symmetry) receive +1 point.

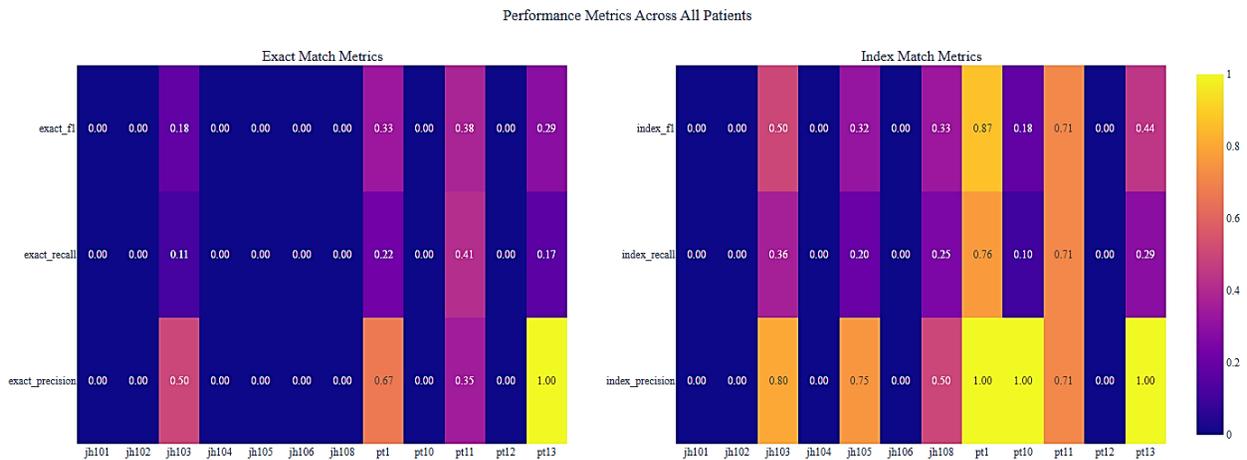

**Fig.6.** Comparative Heatmap Analysis of SOZ Detection Performance: Exact vs. Index Matching Methods Across Patient Cohort: This dual heatmap visualizes SOZ detection performance metrics (precision, recall, F1) across all patients for both exact and index matching methods. Values are color-scaled (viridis colormap) from low (purple) to high (yellow) performance, with exact numerical values overlaid. The left panel shows exact channel matching requiring full SOZ-outlier

correspondence, while the right panel shows index matching allowing partial matches. These heatmaps compare: (1) inter-patient performance variability (vertical patterns), (2) relative metric strengths within patients (horizontal patterns), and (3) differences between methods.

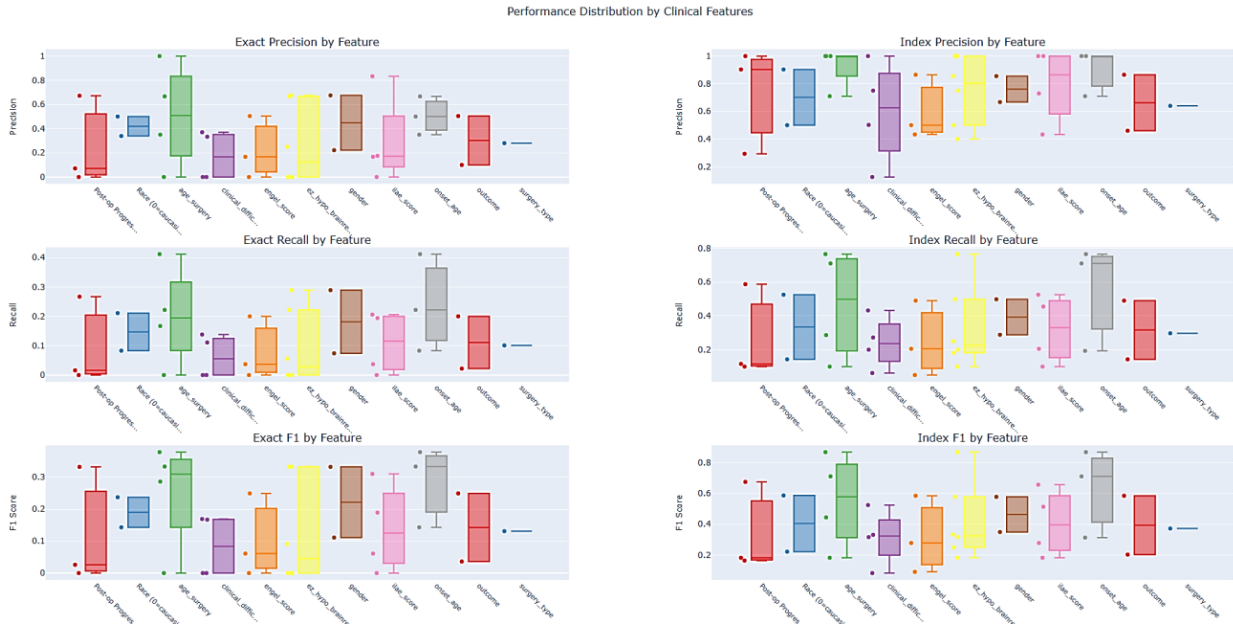

**Fig.7. Comparison of SOZ Detection Performance Metrics between exact matching and index matching techniques:** Box plots showing distributions of precision, recall, and F1 scores for seizure onset zone (SOZ) detection using exact channel matching (left panel) and weighted index matching (right panel) across different clinical phenotypes. Whiskers indicate variability, boxes represent interquartile ranges (IQR), and horizontal lines denote median values. Outliers are shown as individual points. The weighted index matching method incorporates hemispheric weighting, improving detection performance compared to exact matching.

The heatmaps in Fig.6 were generated by aggregating performance metrics (precision, recall, and F1 scores) across all patients for both exact and index match methods. Two adjacent heatmaps compare the performance of exact match (left) and index match (right) methods. The heatmap rows represent performance metrics: Precision (correct SOZ predictions among detected outliers), Recall (true SOZ channels correctly identified), and F1 Score (harmonic mean of precision and recall), while columns correspond to individual patients labeled by their ID. Warmer colors (yellow/orange) indicate higher scores (better performance), cooler colors (blue/purple) denote lower scores, and overlaid numerical values enable precise interpretation of each metric-patient combination. The heatmaps enable direct comparison of method performance across the cohort.

Fig.7 highlights how detection performance varies across clinical phenotypes and demonstrates the relative advantage of index matching that incorporates spatial weighting. This visualization was generated by analyzing patient-specific data using Local Outlier Factor (LOF) anomaly detection on standardized spectral features (start frequency, end frequency, duration) aggregated at the channel level, then comparing the detected outliers against clinically identified seizure onset zones (SOZs) to calculate performance metrics (precision, recall, F1) through both exact channel matching and weighted index matching (accounting for hemispheric location). The boxplot represents the distribution of performance metrics across patients, showing median values, interquartile ranges, and outliers, allowing comparison of SOZ detection accuracy between different clinical characteristics (e.g., Engel scores, surgical types) while controlling for variability in channel counts and SOZ distributions.

According to Table 1, the method demonstrates the strongest performance for "Seizure-free" patients (3 patients), achieving high Index Precision (0.903 ± 0.168) and moderate Index Recall (0.587 ± 0.262). For "S" (Successful outcome, 4 patients), performance is also high, particularly in Index Precision (0.865 ± 0.157). In contrast, "F" (Failure, 5 patients) exhibits low exact-match performance and moderate Index Precision (0.460 ± 0.456). Overall, the method performs best for "Seizure-free" patients and "S" (Successful) outcomes, particularly under index-based matching.

**Table 1. Performance Evaluation by Outcome Category**

| Clinical Feature | Feature Value | Patient Count | Exact Precision | Exact Recall | Exact F1 | Index Precision | Index Recall | Index F1 |
|---|---|---|---|---|---|---|---|---|
| Post-op Progress | 4 GTCs in 1st 6 weeks postop, then seizure free since | 1 | 0.000 | 0.000 | 0.000 | 1.000 | 0.100 | 0.182 |
| | na | 7 | 0.071 ± 0.189 | 0.016 ± 0.042 | 0.026 ± 0.069 | 0.293 ± 0.377 | 0.116 ± 0.153 | 0.164 ± 0.213 |
| | seizure free | 3 | 0.672 ± 0.325 | 0.267 ± 0.129 | 0.332 ± 0.046 | 0.903 ± 0.168 | 0.587 ± 0.262 | 0.674 ± 0.213 |
| | seizure free x 2 years, then recurred after AEDs tapered off | 1 | 0.000 | 0.000 | 0.000 | 0.000 | 0.000 | 0.000 |
| Outcome | F | 5 | 0.100 ± 0.224 | 0.022 ± 0.050 | 0.036 ± 0.081 | 0.460 ± 0.456 | 0.143 ± 0.160 | 0.203 ± 0.217 |
| | NR | 3 | 0.000 ± 0.000 | 0.000 ± 0.000 | 0.000 ± 0.000 | 0.000 ± 0.000 | 0.000 ± 0.000 | 0.000 ± 0.000 |
| | S | 4 | 0.504 ± 0.428 | 0.200 ± 0.170 | 0.249 ± 0.170 | 0.865 ± 0.157 | 0.490 ± 0.288 | 0.584 ± 0.250 |

## 4. Discussion

A 2024 study by Di Giacomo and colleagues demonstrates that chirp patterns detected in stereoelectroencephalography (SEEG) recordings identify the epileptogenic zone (EZ) in patients with drug-resistant focal epilepsy. Analyzing retrospective data from 176 individuals, the researchers observed that chirps—short, high-frequency signals with a progressive frequency shift—were present in more than 95% of seizures characterized by low-voltage fast activity, irrespective of pathological or anatomical differences. Notably, the degree of overlap between chirp locations and visually determined EZs strongly correlated with improved surgical success rates (p < .036). Conversely, mismatches between chirp distribution and conventional EZ mapping were associated with reduced seizure control (p = .01). Patients who achieved complete seizure freedom (Engel class Ia) often had chirp-involved regions resected or treated via thermocoagulation, underscoring their clinical relevance. These results indicate that spectral analysis of chirps can serve as an efficient, semi-automated tool to refine EZ localization, particularly in challenging or MRI-negative cases, thereby improving both diagnostic accuracy and surgical decision-making (Di Giacomo et al., 2024). Though this research has established that chirps can be used as markers of epileptogenic tissue and surgical efficacy, a systematic examination of their spatiotemporal characteristics and clinical utility is still lacking.

Our approach was quantifying chirp characteristics (Onset frequency, offset frequency, temporal duration) and applying anomaly detection to identify outlier channels that might correspond to pathological activity. The use of the Local Outlier Factor (LOF) algorithm extended prior work by offering a semi-automated method to detect deviations in chirp features, reducing reliance on purely visual EZ delineation.

### 4.1. Spatial Convergence Between Computational and Clinical Annotations

Our results demonstrated that LOF-detected outliers exhibited spatial overlap with clinically annotated SOZ channels (Figs. 4–6). This was consistent with Di Giacomo et al.'s observation that chirp-EZ alignment predicted surgical success. Notably, index matching—which incorporated hemispheric weighting—outperformed exact matching in precision and recall (Fig. 7), suggesting that spatial proximity enhanced SOZ detection accuracy. This finding supported the hypothesis that epileptogenic activity was not only electrophysiologically distinct but also spatially clustered, reinforcing the utility of computational approaches in refining EZ localization.

### 4.2. Clinical Relevance and Outcome Correlation

The strongest performance was observed in seizure-free patients, where index precision reached $0.903 \pm 0.168$ (Table 1). This aligned with Di Giacomo et al.'s finding that resection of chirp-involved regions correlated with favorable outcomes. Conversely, poorer performance in non-responders ("F" group) suggested that mismatches between computational and clinical SOZ definitions might reflect incomplete EZ resection.

**Acknowledgments**

The authors gratefully acknowledge the financial support of the Canadian Neuroanalytics Scholars Program and The Hilary & Galen Weston Foundation. We also recognize the valuable support provided by Campus Alberta Neuroscience, the Hotchkiss Brain Institute at the University of Calgary, the Ontario Brain Institute, and the Neuro at McGill University. Their contributions were essential to the advancement of this research project.

**Appendix: Visual Evidence of SOZ Concordance Using Chirp-Derived Spatial Outliers**

Fig.4 (Supp1) – Fig.4 (Supp4) demonstrate that 3D chirp embeddings highlight SOZ-associated channels as spatial outliers, showing concordance with clinical annotations. Further supporting this alignment, Fig.4 (Supp5) – Fig.4 (Supp7) present radial embeddings based on onset/offset frequency and chirp duration, revealing distinct outlier clusters that correspond to SOZ contacts.

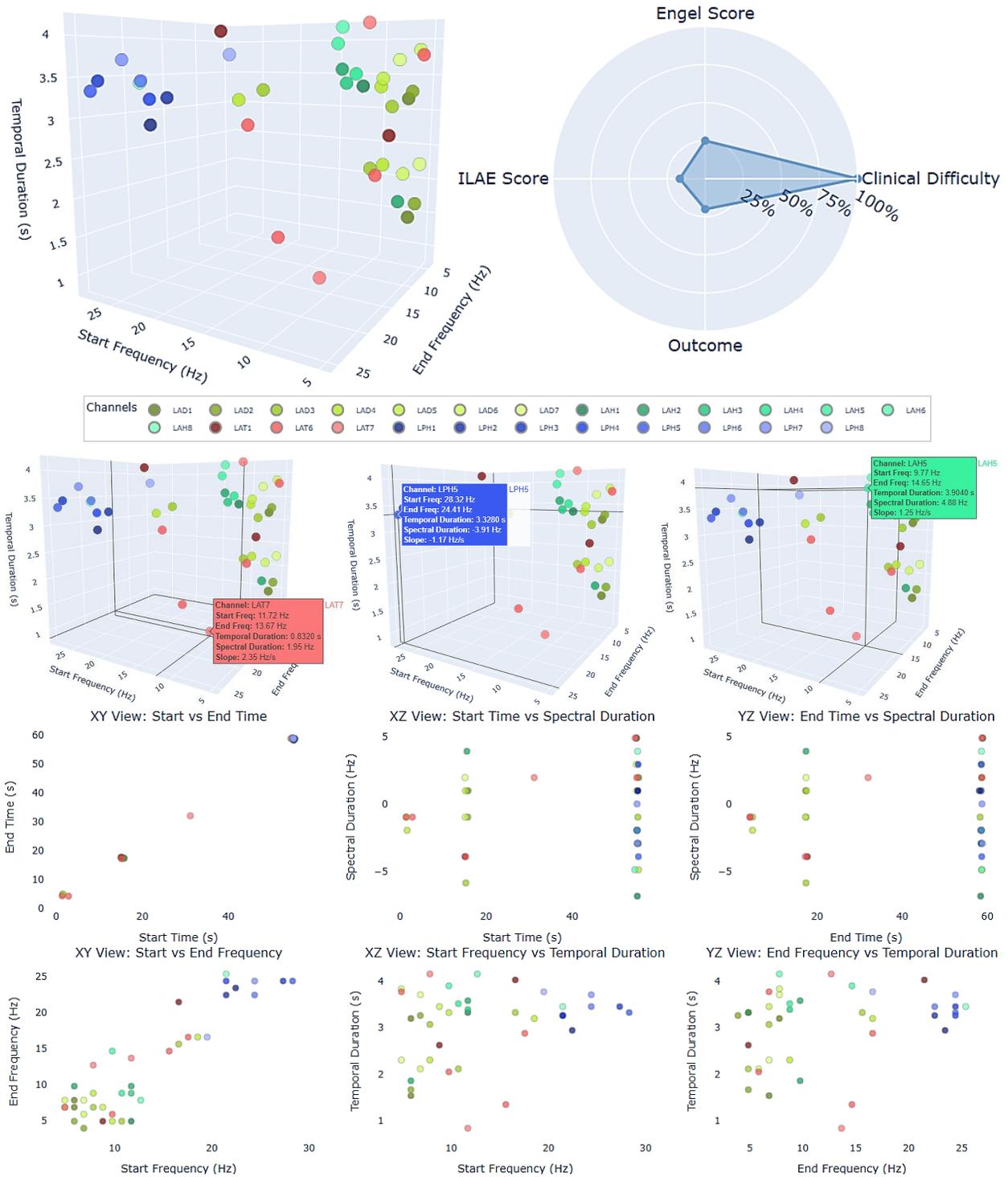

**Fig.4 (Supp1). Embedding of Patient jh101 Chirp Characteristics in a 3D Feature Space:** The embedding is constructed using three key features extracted from individual chirps: temporal duration, onset frequency, and offset frequency. Each chirp is represented as a single point in a three-dimensional feature space, with its coordinates corresponding to these attributes. Thus, for each recording channel, chirp characteristics are encoded as a 3D vector of the form [duration,

onset frequency, offset frequency]. This low-dimensional representation facilitates the spatial comparison of chirp features across channels and enables direct visualization of their distribution. Notably, certain channels, such as LAT7, appear spatially isolated within this feature space, indicating that their chirp profiles are distinct from those of other channels. This approach supports the identification of potential outliers and clustering patterns based on chirp dynamics.

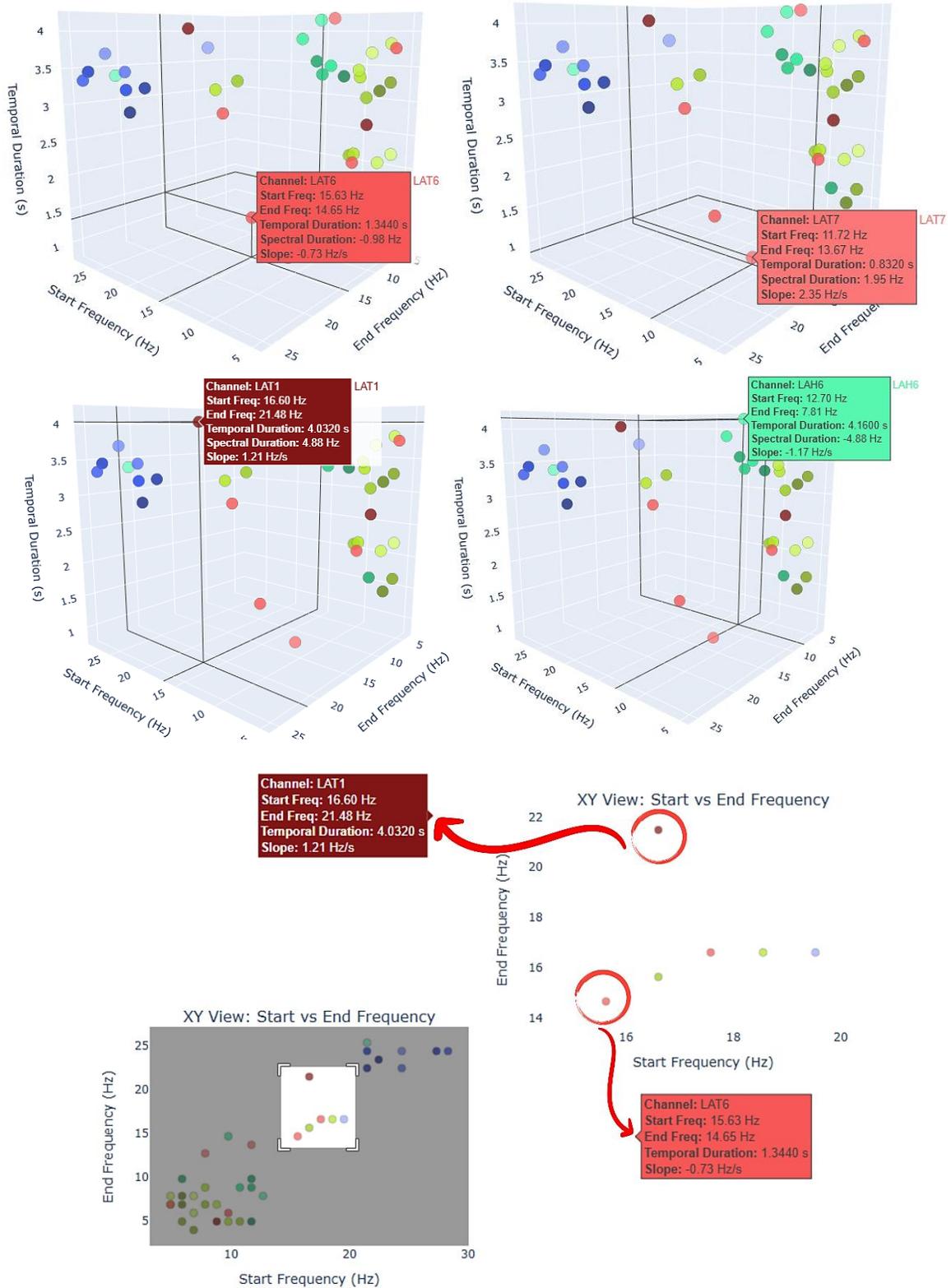

**Fig.4 (Supp2). Comparison of 3D Feature Embedding with Clinical Annotations (Patient jh101):** A zoomed view of Fig.4 reveals a notable correspondence between the 3D feature embedding of chirp characteristics and the clinically annotated seizure onset zone (SOZ) as identified by the neurosurgical team. Specifically, channels labeled by the surgeon as part of the SOZ tend to cluster within a distinct subregion of the 3D feature space. According to clinical reports, the epileptogenic zone (EZ) is localized to the left anterior temporal lobe and the left amygdala, with relative sparing of the hippocampus. The SOZ includes electrode contacts LAT1–2, LAT6–7, and LAH6. The alignment between the data-driven feature space and clinical annotations suggests that chirp characteristics may reflect underlying epileptogenic dynamics.

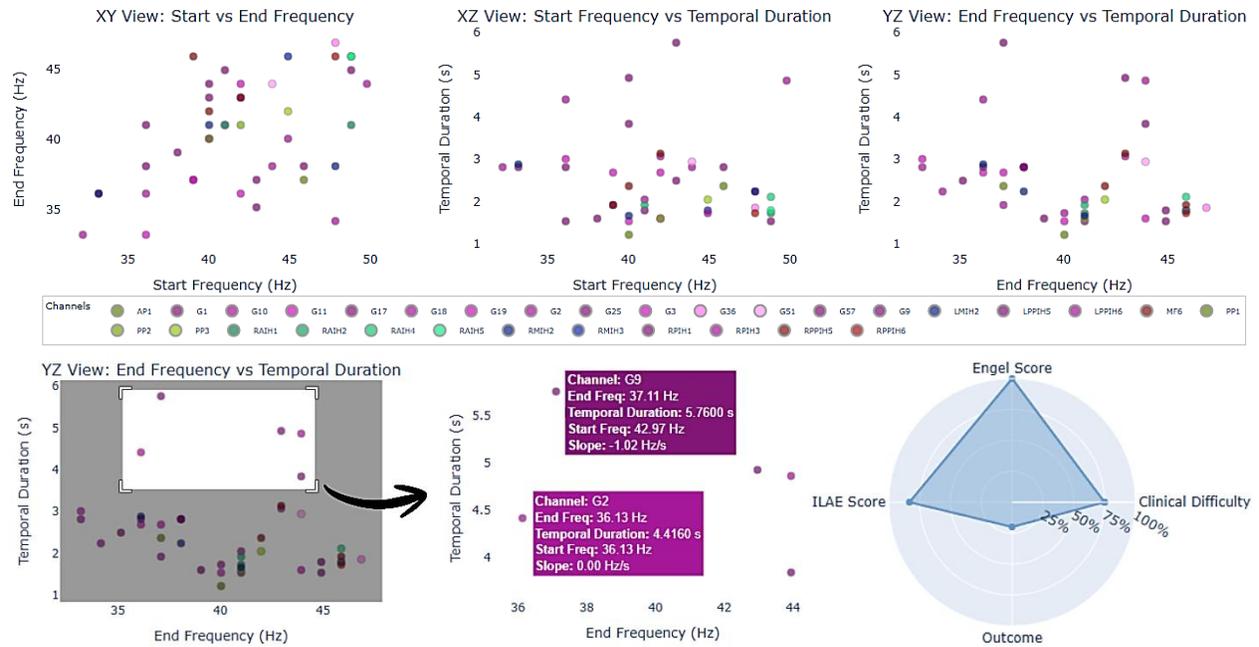

**Fig.4 (Supp3). Alignment of Feature Embedding Outliers with Clinically Annotated SOZ in Patient pt13:** A focused examination of the feature embedding for Patient pt13 reveals that outlier channels within the 3D chirp characteristic space align with the seizure onset zone (SOZ) as identified through clinical annotations by the surgical team. According to these annotations, the SOZ is localized to the right parietal lobe, with involved contacts including G1–2, G9–10, and G17–18. Notably, the patient achieved seizure freedom postoperatively, suggesting that the identified SOZ was accurately targeted. The spatial alignment between embedding outliers and clinically defined SOZ regions supports the potential utility of chirp-based features in localizing epileptogenic tissue.

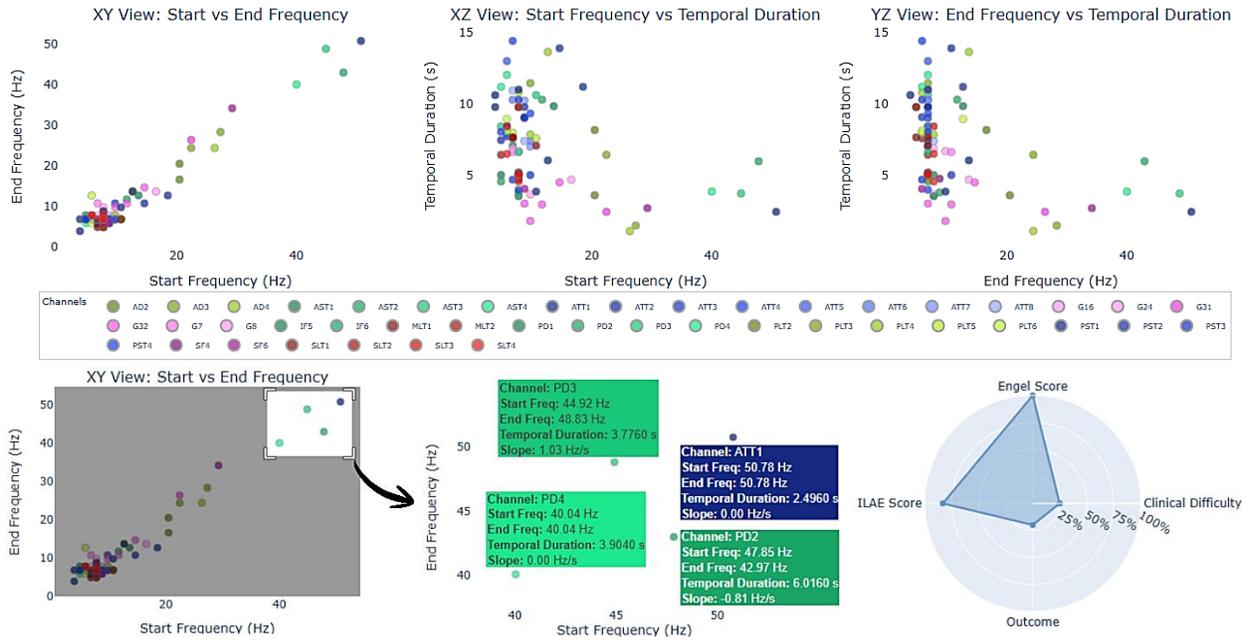

**Fig.4 (Supp4). Outlier Positioning of Clinically Annotated SOZ Channels in Patient pt1's Feature Embedding:** A zoomed-in analysis of the 3D feature space for Patient pt1 demonstrates that channels clinically annotated as part of the seizure onset zone (SOZ) are positioned as outliers relative to the distribution of other channels. These SOZ-designated channels occupy distinct regions within the chirp-based feature space, suggesting divergence in signal characteristics. According to clinical annotations, the SOZ is localized to the right anterior temporal lobe, involving contacts PD1–4, AD1–4, and ATT1–2. The patient achieved seizure freedom following intervention, further supporting the accuracy of the SOZ localization and the relevance of the identified feature-based outliers.

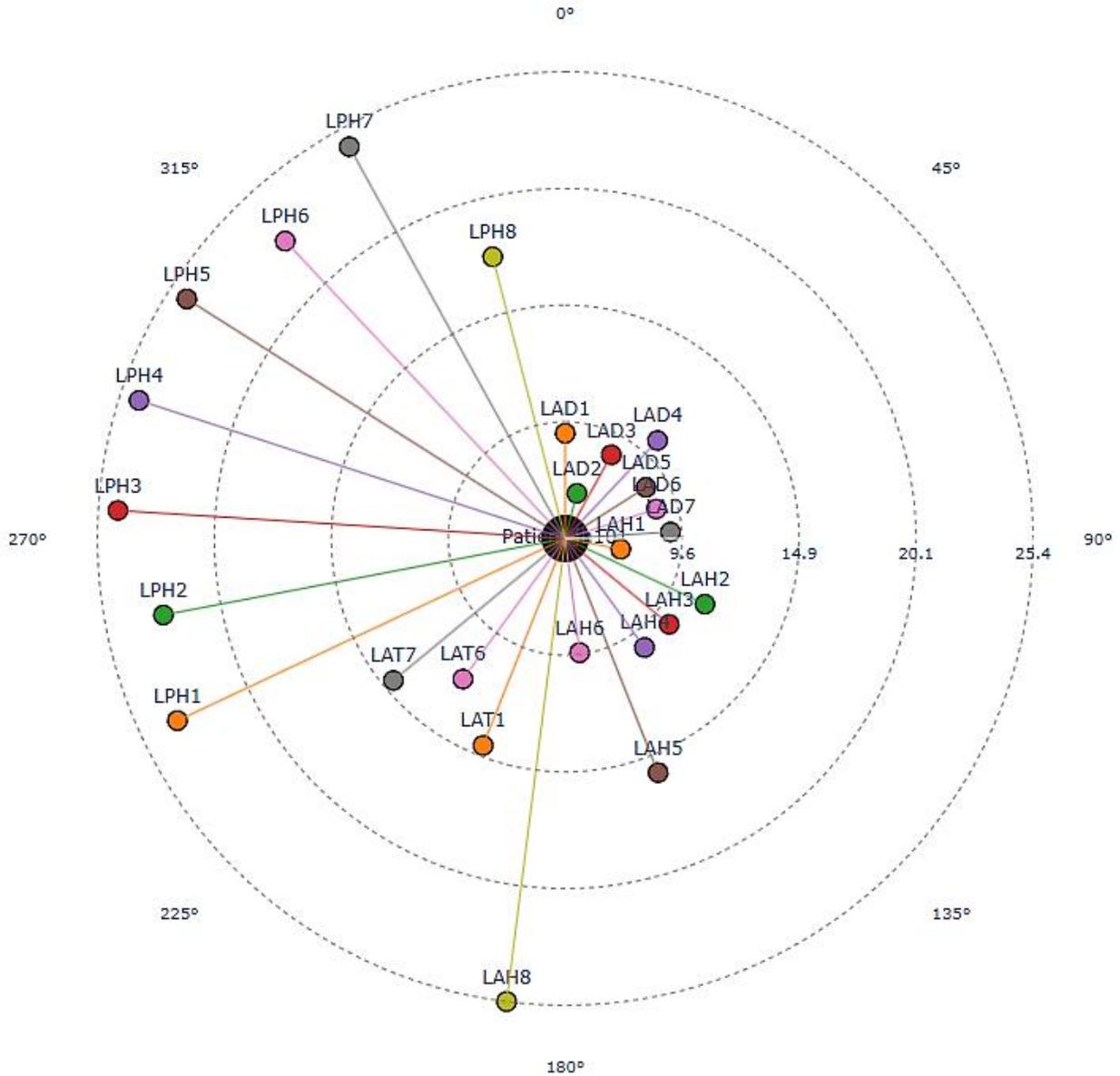

**Fig.4 (Supp5). Radial End Frequency Distribution Reveals Outlier Cluster Corresponding to Clinically Annotated SOZ in Patient jh101:** Analysis of the distribution of channels based on Radial End Frequency (Hz) (Chirp offset frequency in radial space) demonstrates that channels with similar chirp frequency offset tend to cluster at comparable radial distances. Specifically, a large number of channels are concentrated around a radius of approximately 9.6 Hz, while another substantial group is observed near 25.4 Hz. A smaller subset of channels appears around 14.9 Hz, forming a distinct cluster. These channels at ~14.9 Hz (LAT1 and LAT6-7) deviate from the two dominant groupings and appear as outliers in the radial distribution. When compared with clinical annotations, this outlier group corresponds to channels identified by the surgical team as part of the seizure onset zone (SOZ). For patient jh101, the epileptogenic zone (EZ) was localized to the left anterior temporal lobe and left amygdala, with relative preservation of the hippocampus. SOZ contacts include LAT1–2, LAT6–7, and LAH6.

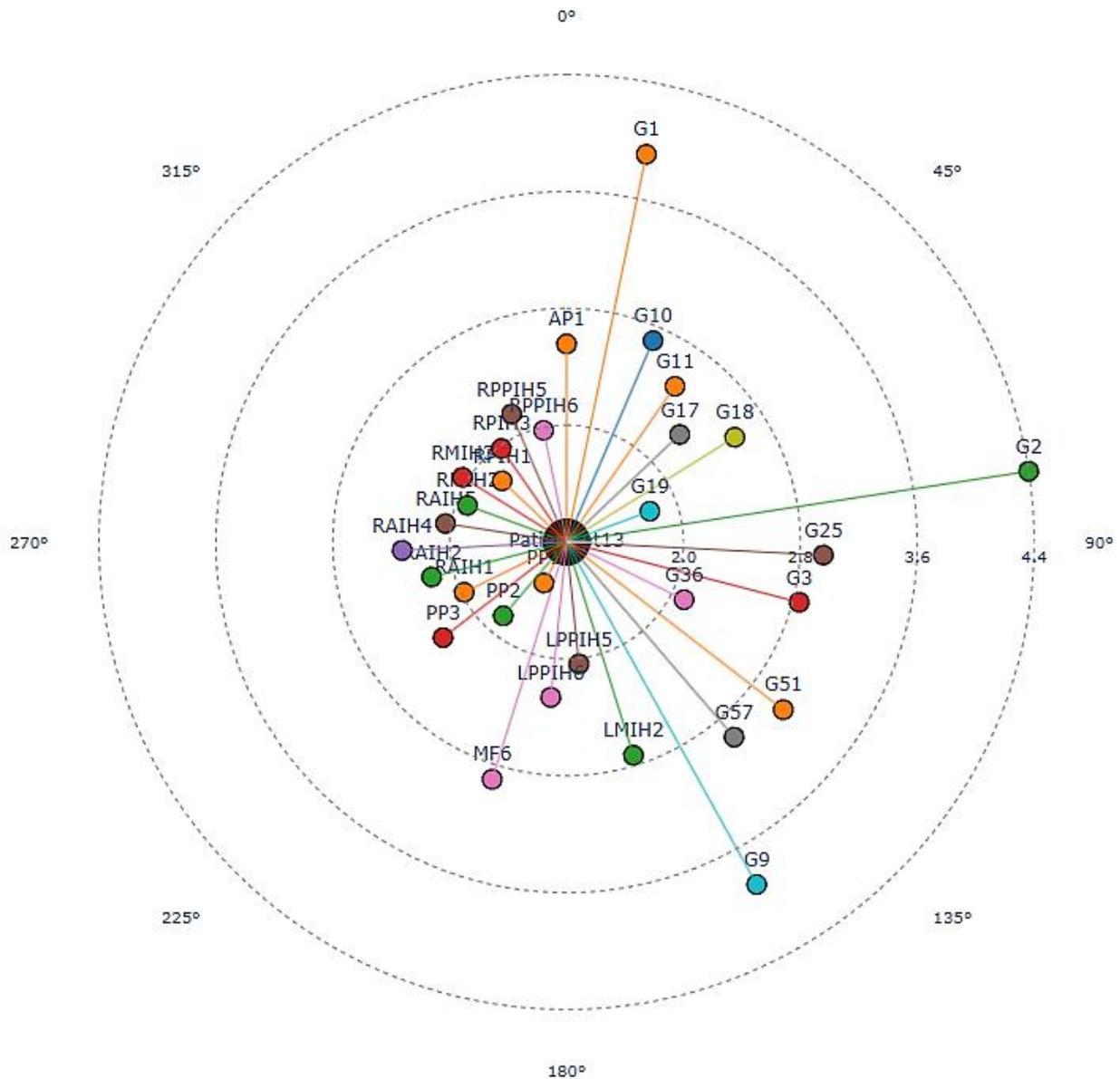

**Fig.4 (Supp6). Radial Duration Distribution Identifies SOZ-Associated Outliers in Patient pt13**

In Patient pt13, analysis of the Radial Duration (s) distribution (Chirp duration distribution in radial space) reveals that most channels are clustered around or below 2.8 seconds, representing the dominant group within the feature space. In contrast, a subset of channels, specifically G1, G2, and G9, exhibit significantly longer chirp durations, exceeding 3.6 seconds, and are positioned as outliers relative to the main distribution. Notably, these outlier channels correspond to electrode contacts clinically annotated as part of the seizure onset zone (SOZ) by the surgical team. According to clinical documentation, the SOZ is localized to the right parietal lobe and includes contacts G1–2, G9–10, and G17–18. The patient achieved seizure freedom following intervention, further supporting the clinical relevance of these channels. In this patient, the alignment was observed between elevated chirp durations and SOZ annotations.

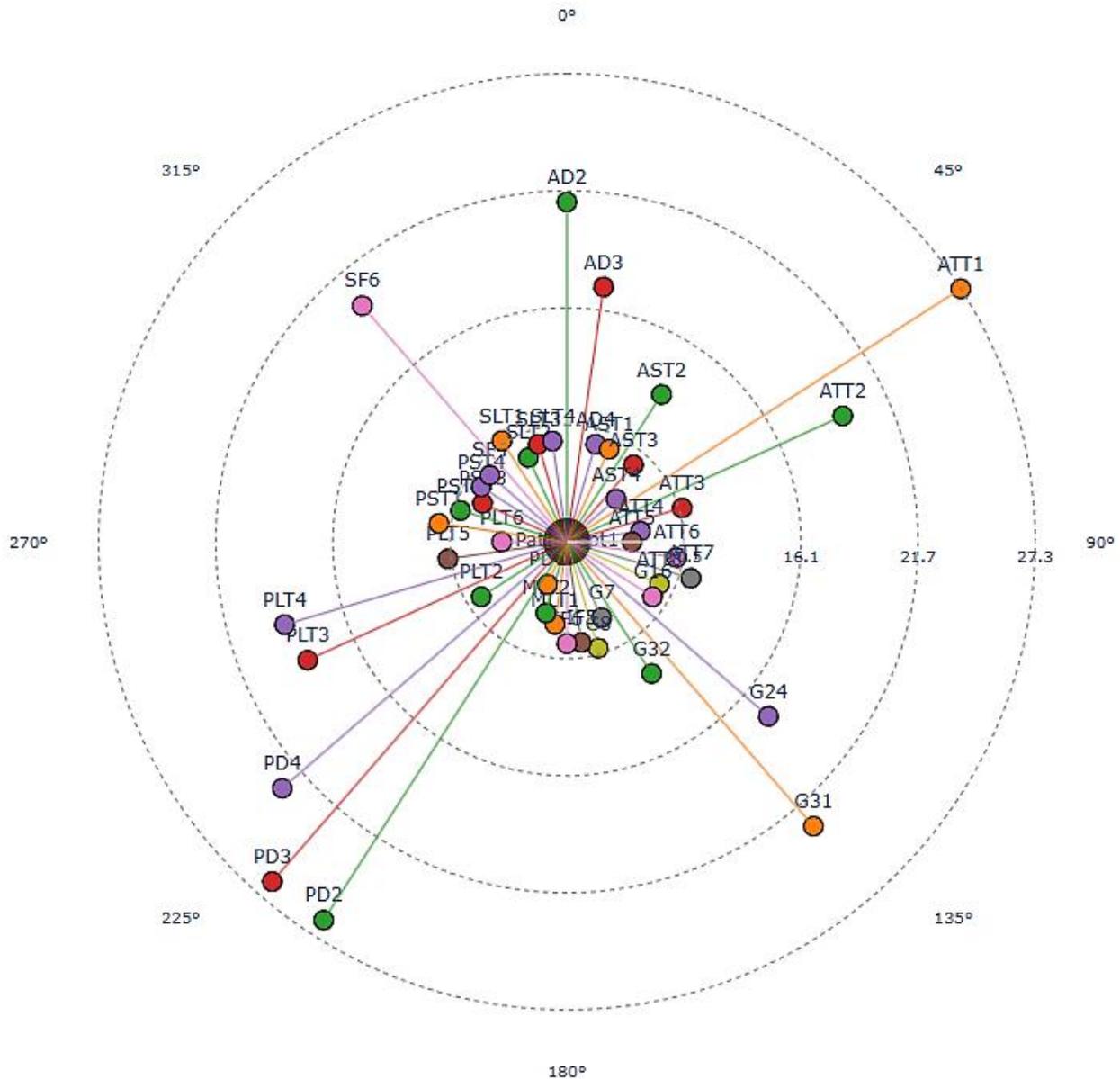

**Fig.4 (Supp7). Radial Start Frequency Reveals SOZ-Associated Outliers in Patient pt1**
In Patient pt1, analysis of Radial Start Frequency (Hz), reflecting chirp onset frequency, demonstrates separation of channels within the radial feature space. Most channels are concentrated within lower onset frequency radii, whereas a distinct set of outliers, including ATT1, ATT2, AD2, AD3, PD2, PD3, and PD4, are positioned at radii exceeding 16.1 Hz. Notably, these outlier channels correspond to electrode contacts clinically annotated as part of the seizure onset zone (SOZ) by the surgical team. According to clinical reports, the SOZ is localized to the right anterior temporal lobe and includes contacts PD1–4, AD1–4, and ATT1–2. The patient achieved seizure freedom following surgical intervention, supporting the accuracy of SOZ identification. In this patient, the alignment was observed between high chirp onset frequencies and clinically defined SOZ contacts.